%% file: main.tex
\documentclass[conference,9pt]{IEEEtran}
\IEEEoverridecommandlockouts

\usepackage[T1]{fontenc}
\usepackage[utf8]{inputenc}
\usepackage{amsmath,amssymb}
\usepackage{algorithm}
\usepackage[noend]{algpseudocode}
\usepackage{listings}
\usepackage{booktabs,array}
\usepackage{makecell}
\usepackage{graphicx}
\usepackage{xcolor}
\usepackage{cite}
\usepackage{url}
\usepackage[hidelinks]{hyperref}
\usepackage{xurl}
\usepackage{xspace}

\newcommand{\fra}{FRA\xspace}
\newcommand{\afra}{AFRA\xspace}
\newcommand{\jasper}{Jasper\xspace}

\newtheorem{prop}{Proposition}

\definecolor{ours}{HTML}{0072B2}
\newcommand{\ours}[1]{\textcolor{ours}{#1}}

\newcommand{\mstd}[2]{$#1\,{\scriptstyle\pm #2}$}

\lstdefinelanguage{SVA}{sensitive=true, alsoletter={\$},
  morekeywords={$past,$stable,$onehot,$onehot0,$countones,$rose,$fell,$changed,$isunknown},
  morekeywords=[2]{cover,property,assert,assume,disable,iff,sequence}, morecomment=[l]{//},
  literate={|->}{{\textcolor{ours}{|->}}}3 {|=>}{{\textcolor{ours}{|=>}}}3 {\#\#}{{\textcolor{ours}{\#\#}}}2 {-}{{-}}1 {*}{{*}}1}
\lstdefinestyle{sva}{language=SVA, basicstyle=\fontfamily{zi4}\selectfont\scriptsize, columns=fullflexible,
  keepspaces=true, upquote=true, keywordstyle=\color{ours}, keywordstyle=[2]\bfseries,
  commentstyle=\color{black!55}, numbers=left, numberstyle=\scriptsize, numbersep=6pt,
  xleftmargin=1.5em, aboveskip=2pt, belowskip=0pt, escapechar=@}

\begin{document}

\title{From Bounded to Unbounded RTL Proofs via Agentic Construction of Helper Assertions}

\author{\IEEEauthorblockN{Zijian Ding\IEEEauthorrefmark{1}\IEEEauthorrefmark{3}\thanks{\IEEEauthorrefmark{3}Work done during an internship at NVIDIA.}, Yunsheng Bai\IEEEauthorrefmark{2}, Ghaith Bany Hamad\IEEEauthorrefmark{2}, Chenhui Deng\IEEEauthorrefmark{2}, Cunxi Yu\IEEEauthorrefmark{2}, Jason Cong\IEEEauthorrefmark{1}, Brucek Khailany\IEEEauthorrefmark{2}}
\IEEEauthorblockA{\IEEEauthorrefmark{1}UCLA\qquad \IEEEauthorrefmark{2}NVIDIA\\
\{bradyd, cong\}@cs.ucla.edu, \{yunshengb, gbanyhamad, cdeng, cunxiy, bkhailany\}@nvidia.com}}

\maketitle

\begin{abstract}
Constructing helper assertions is a major challenge in scaling formal verification (FV) to complex RTL designs. Target properties can remain unresolved even if running the engine for months, and recent LLM-based approaches have had limited success on large designs. We study this challenge through proof graphs that capture dependencies among target properties and helper assertions. Our analysis identifies three recurring failure modes of LLM agents. To address them, we combine behavioral modeling and spurious-loop search to construct proofs with high depth and dense connectivity. We further define a progress measure on the proof graph and use it as a dense reward to train a customized model. Across 587 target properties spanning 13 design families, our approach proves 80.5\% of targets on average across families, compared with 36.4\% for the strongest baseline. It is also the only evaluated method to prove all 104 targets of a 68.4K-line production design. On production targets, helper-strengthened proofs have a median checking time of 7.3 minutes, while the proof engine alone leaves these targets unresolved after 12 hours.
\end{abstract}

\begin{IEEEkeywords}
formal property verification, proof closure, LLM agents, language models, verification sign-off
\end{IEEEkeywords}

\input{sections/intro}
\input{sections/prelim}
\input{sections/method}

\input{sections/eval}
\input{sections/related}
\input{sections/disclosure}
\input{sections/conc}

\bibliographystyle{IEEEtran}
\bibliography{references}

\clearpage
\raggedbottom
\input{sections/appendix}

\end{document}

%% file: sections/intro.tex
\section{Introduction}
\label{sec:intro}
\suppressfloats[t]

As modern system-on-chip (SoC) and accelerator architectures scale in concurrency and pipeline depth, functional verification has become the primary bottleneck in silicon development~\cite{seligman2015formal}. Simulation samples only a fraction of reachable states, so deep-cycle bugs such as credit deadlocks can escape into silicon and cause costly respins~\cite{wilson2024,roy2025hierarchical}. Formal verification (FV) via model checking~\cite{biere1999symbolic,bradley2011sat} holds the promise of \emph{full proof closure}: proving that a property holds on all input sequences for unbounded time.

\begin{figure}[t]
  \centering
  \includegraphics[width=\columnwidth]{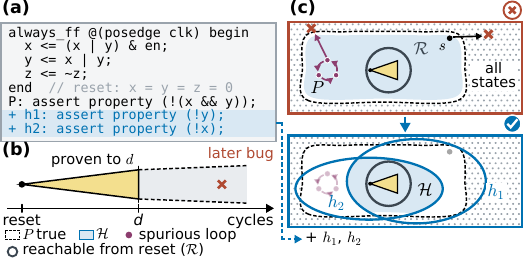}
  \caption{Bounded versus proof closure. 
  (a) A 3-bit design with its target $P$ and the appended helpers $h_1$ and $h_2$. (b) A bounded proof of $P$ covers $d$ cycles from reset and can miss a later bug. (c) Assuming only $P$, the consecution check fails with CTIs (Sec. \ref{sec:fra-prelim}), as states outside the reachable states $\mathcal{R}$ step out of $P$. With $h_1$ and $h_2$ assumed, these states are excluded and $P$ is proven.
  }
  \label{fig:teaser}
\end{figure}

\begin{figure*}[t]
  \centering
  \includegraphics[width=0.287\textwidth]{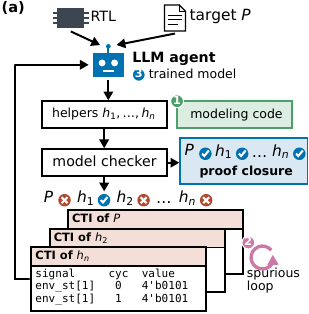}\hfill
  \includegraphics[width=0.7\textwidth]{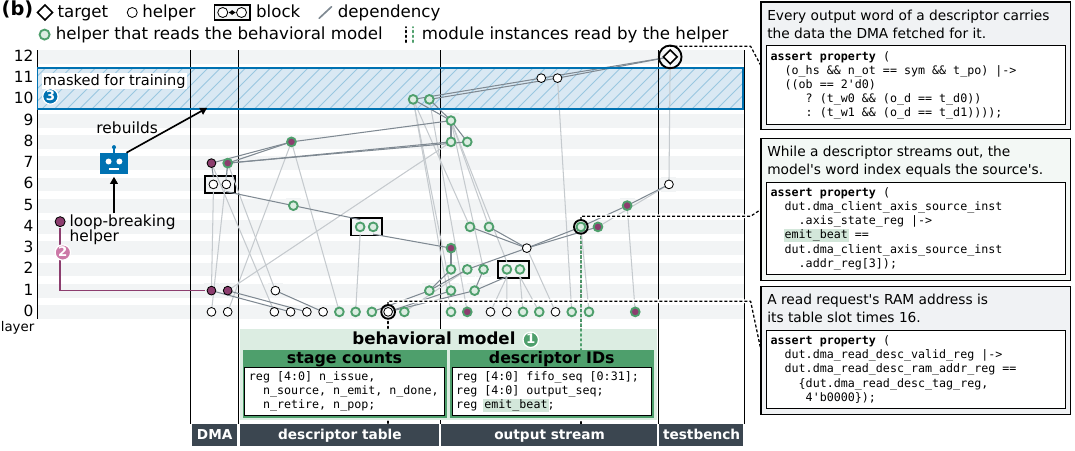}
  \caption{(a) The basic flow of an LLM agent for proof closure, and the three components that \afra introduces, behavioral modeling code (Sec. \ref{sec:modeling}), spurious-loop search (Sec. \ref{sec:witness}) and a trained model (Sec. \ref{sec:train}). Proof closure is reached only when every helper and the target are proven. (b) Post-closure analysis on the dependency graph of \afra's proof for the descriptor-fetch unit of the open-source Corundum NIC~\cite{forencich2020corundum}, where each edge shows a necessary dependency (Sec. \ref{sec:graph}).
  }
  \label{fig:anchor}
\end{figure*}

Yet on complex RTL blocks, running a formal engine~\cite{synopsysVCFormal,cadence2026jasper} on a hard property often ends in a \emph{bounded proof}: no counterexample within $d$ cycles of reset, while a bug at depth $d{+}1$ can still escape (Fig. \ref{fig:teaser}(b)). Fully proving such a property requires design-specific abstractions and missing invariants~\cite{burch1994automatic,mcmillan1998verification,agrawal2020,hernandez2025}, which can take experienced FV engineers months~\cite{reid2016end,agrawal2020,yeung2022raising,roy2025hierarchical}.

Recent works explore large language models (LLMs) to automate invariant discovery~\cite{peled2025large,su2026cill}. A typical basic agent flow (Fig. \ref{fig:anchor}(a)) takes a design and target properties (targets), lets an LLM propose helper assertions (helpers) written in SVA, receives consecution counterexamples (CTIs) as feedback from a model checker for every helper and the target, and lets the LLM revise the helper set. Proof closure is achieved only when every helper and the target are proven. It is not surprising that this approach can be very effective, especially on properties where existing model checking engines fall short~\cite{gilday2021human}. On our test set, a frontier coding agent can construct proofs for targets that a commercial formal engine cannot fully prove, with up to $454\times$ speedup in proof checking (Fig. \ref{fig:prooftime}). However, constructing a helper proof with an agent is slow and costly, and when the agent fails, it is hard to analyze the failure modes, especially since manually constructing helpers is hard for any existing method and even for human experts. This leads us to a central question:

\begin{center}
\emph{How to systematically develop SVA helpers (with agents) that lead to proof closure, on properties where existing methods fall short?}
\end{center}

To see where the basic flow fails, we do a post-closure analysis of the proofs constructed by agents. We observe that a helper-strengthened proof forms a \textbf{\textit{dependency graph}} (Fig. \ref{fig:anchor}(b)), with local helpers at the bottom and the targets at the top. An edge $h_i\rightarrow h_j$ means that $h_j$ fails its engine check without $h_i$ (Sec. \ref{sec:graph}). On many such graphs, we find that the basic flow fails in three ways. (1) \emph{Depth}: the upper layers relate state spread across module instances, making them hard to express with existing design signals. (2) \emph{Buried root cause}: the connectivity of the graph can be high, so one missing helper makes every helper that depends on it fail, and the root cause is buried in a flood of CTIs that are hard to analyze. (3) \emph{Sparse progress feedback}: the targets are proven only once the helper set is complete, and a larger helper set is not necessarily closer to proof closure. For a naive agent, each round takes over an hour with slow agent generation and proof engine calls, and most rounds end without any new target proven.

To mitigate these challenges, we present \afra (Agentic Full Reset Abstraction), an agentic proof-construction framework that addresses each of these failure modes, with closed models or with an open model that we train. To address depth, we propose \emph{SVA behavioral modeling}, which lets the agent write modeling code for the state the upper layers need. For the root cause, we find that under a fixed window of $K$ cycles, each CTI points to a missing dependency. We propose a novel two-phase search. The agent first breaks the spurious loops, which persist for any $K$ and point to the dependencies the targets are missing, and then chooses the window $K$ for the remaining CTIs. We describe the window in Sec. \ref{sec:fra-prelim} and spurious loops in Sec.~\ref{sec:witness}. For sparse progress feedback, we propose a \emph{progress measure} on the graph and use it as a dense reward to train a specialized
model on a \emph{curriculum} built from the graph. Together, these three novel components let the agent construct more complex proofs, with more dependencies and, as we show empirically, deeper layers. We summarize our contributions:

\begin{itemize}
  \item \textbf{Agent harness.} We propose behavioral modeling code, with which the agent builds deep proofs, and spurious-loop search, which points the agent to the dependencies a target is missing rather than to a flood of CTIs.
  \item \textbf{Training.} We define a measure of proof progress on the dependency graph of a golden proof, and use it as a dense reward, with a curriculum of partial golden proofs, to train a customized model.
  \item \textbf{Benchmark and evaluation.} On 343 hard properties across 5 production design families and 244 from 8 open-source design families, \afra raises the targets GPT-6.1-sol proves by 44.1 percentage points over the second-best method, and it is the only method that proves all 104 targets of a 68.4K-line production design. With our trained 27B model, \afra improves the solve rate by 9.9 percentage points over the untrained model under the same harness. AFRA-constructed proof check is up to 454$\times$ faster than commercial engine.
\end{itemize}

%% file: sections/prelim.tex
\section{Preliminaries}
\label{sec:prelim}
Given fixed RTL with its reset and environment assumptions, a target safety property $P$\footnote{We consider safety properties, whose violations admit finite witnesses.}, and possible additional context such as specification, formal property verification (FPV) asks whether $P$ holds on the states $\mathcal{R}\subseteq\mathcal{S}$ reachable from reset under the assumptions, where $\mathcal{S}$ is the set of all design states. A formal engine~\cite{cadence2026jasper,synopsysVCFormal} returns one of three results. \emph{Proven} means $P$ holds on every reachable state for all time, \emph{falsified} means a trace from reset violates $P$, and \emph{bounded} means the engine finds no violation within $d$ cycles of reset before its budget runs out, which the engine reports as \emph{undetermined}. 
Bounded model checking (BMC)~\cite{biere1999symbolic} checks $P$ only within $d$ cycles of reset, so it can return a falsified or a bounded result, but never a proven one.

\label{sec:fra-prelim}
An \textbf{\emph{invariant}} is a property that holds in every state reachable from reset.
Full reset abstraction (\fra) checks two obligations: (i)~a \emph{base case} that an assertion holds initially at reset, and (ii)~a \emph{\textbf{consecution}} (induction) check~\cite{bradley2011sat}: from \emph{any} state satisfying the assertion (register reset values released), one legal transition preserves it.
A true target $P$ often fails \textbf{consecution} because $P$ alone allows unreachable start states. For example, in the design of Fig. \ref{fig:teaser}(a), $P$ holds on every reachable state, yet its consecution check fails from the unreachable state $xyz=010$. When $\mathit{en}=1$, both $x$ and $y$ become 1, so $P$ fails.
\textbf{Helper assertions} are further invariants, proven alongside $P$, that exclude such states. The helpers $h_1$ and $h_2$ of Fig. \ref{fig:teaser}(a) rule out every state that leads to such a failure, completing the proof.
For a true $P$ a strengthening always exists, but none is guaranteed to be small or easy to find~\cite{su2026cill,hu2026enhancing}.

Let $\mathcal{H}=\{P,h_1,\dots,h_m\}$ denote a set of properties including the target property and helper assertions. With a \emph{window} of $K\ge1$ cycles, a formal engine can assume $\mathcal{H}$ on $K$ consecutive states $s_1,\dots,s_K$ before checking it on the next state $s_{K+1}$, and the base case extends to $K$ cycles from reset~\cite{sheeran2000checking}. A path from an arbitrary state that stays in $\mathcal{H}$ for $K$ cycles and exits at step $K{+}1$ is a CTI~\cite{bradley2007checking}.

%% file: sections/method.tex
\section{Methodology}
\label{sec:method}

\subsection{Basic Agent Flow}
\label{sec:flow}

The LLM agent receives the input and creates and edits the property set $\mathcal{H}$, checks it with the formal engine provided to it, and iterates within a total time budget ($T$). One \emph{engine call}, such as the command ``\texttt{prove -sst}'' in \jasper{}, checks $\mathcal{H}$ together. It runs the base case from reset and the consecution check with every property assumed, and reports each property as proven, falsified by a trace from reset, or unproven with a CTI. We then provide the status of each helper and target and the CTIs to the agent by writing them to its workspace as TSV files. We give a short time limit ($T_1$) for the engine call during the run, and observe that the agent's reasoning usually takes much longer than the engine run. The run stops either at proof closure, when every property in $\mathcal{H}$ is proven, or when the time budget runs out. We then check the agent's final property set again with a time limit ($T_2$), and count the targets this final evaluation proves.

\textbf{Soundness.} The soundness of the flow can be stated as follows. (1) Proof closure can never be achieved if any helper is false, or if the target is false, with respect to the input RTL. (2) Undetermined cases (a timeout after the agent run) never decrease the sign-off confidence, since the tool still gives a bounded proof. (3) We count a target only when the final evaluation proves it and every helper it assumes, so a partial result never rests on an unproven helper. We further use a fixed evaluation flow that only takes in $\mathcal{H}$, which is an agent-written SVA file, checks that it adds no extra assumptions, and binds it to a fresh copy of the input task, to ensure no possibility of modifying the source code or scripts. The proof of (1) is well-established in classic literature~\cite{sheeran2000checking,bjesse2000sat}, and we omit it here.

\subsection{Proof Dependency Graph}
\label{sec:graph}

We start by analyzing the limitations of the basic agent flow to motivate our contributions, beginning with the structure of the proofs after closure. A property set $\mathcal{H}$ that fully proves its targets passes one consecution check with all of its properties proven together. However, the set is not flat but forms a directed graph. We define the dependency between properties as follows. We say $h_j$ \textbf{\emph{depends on}} $h_i$, or $h_i$ is \emph{necessary} for $h_j$, when removing $h_i$ from $\mathcal{H}$ and re-checking the rest until a fixed point leaves the consecution check of $h_j$ failing (a CTI), and the \textbf{\emph{proof dependency graph}} has an edge $h_i\rightarrow h_j$ for every such pair, which takes $|\mathcal{H}|$ deletions. We remove each $h_i$, re-check the rest until a fixed point, and add an edge $h_i\rightarrow h_j$ for every $h_j$ that fails, which takes $|\mathcal{H}|$ deletions. We further define the layer of a helper as its depth in the DAG obtained by merging the strongly connected components (SCC) (\emph{blocks}) of the proof dependency graph (Fig. \ref{fig:anchor}).

Upon profiling the proof dependency graphs of many proofs constructed by agents, we find that (1) most proofs are not flat and have high depth (Fig. \ref{fig:graphs}), (2) helpers at deeper layers usually relate modules across the design (Sec. \ref{sec:modeling}), and (3) helpers form complex dependencies. Each CTI (under a fixed window $K$) points to a dependency, but there is no guarantee that the tool provides the dependencies efficiently (e.g., each CTI pointing to a different missing helper), making the root cause analysis challenging for the agent, especially when $\mathcal{H}$ grows larger (Sec. \ref{sec:case}).

We therefore propose behavioral modeling code (Sec. \ref{sec:modeling}) and spurious-loop search (Sec. \ref{sec:witness}), which let the agent build complex full proofs. The graph of a golden proof also gives the progress measure that trains our open model (Sec. \ref{sec:train}).

\subsection{Building Deep Proofs with Behavioral Modeling Code}
\label{sec:modeling}

The helpers in the upper layers of a proof tend to spread across design modules. Across 389 proofs that agents found on our tasks, 59\% of the upper-layer helpers relate several module instances, compared with 35\% in the lower layers. Such a helper needs information that no single module keeps, so it is hard to state with design signals alone. To address this issue, we let the agent write an \textbf{auxiliary behavioral model}, which is SVA modeling code with additional registers and update logic that record this information, such as the history of past inputs, counts, and the identity and order of packets. We instruct the agent to reuse state the design already keeps and to add registers only when the needed information is missing. Asking the agent to construct such a model is still sound, because we refuse any code that drives or constrains the RTL, and every helper must still be proven by the formal engine, so the model can make a proof easier but never unsound. For example, in \afra's proof of Fig. \ref{fig:anchor}, the model counts the descriptors at each stage and records their IDs.

\subsection{Building Densely Connected Proofs with Spurious Loops}
\label{sec:witness}

As Sec. \ref{sec:graph} shows, the root cause of a consecution failure is hidden among many raw CTIs. To find the root cause, we instead search for spurious loops and show them to the agent.

CTIs fall into two kinds (Fig. \ref{fig:trap}). If we trace back the predecessors of an
unreachable state that violates the property, we either reach the start of a chain or enter a \textbf{spurious
loop}, a cycle of unreachable states inside $\mathcal{H}$ with a short exit to a state that violates the property
(Fig. \ref{fig:teaser}(c)). With window $K$, we assume $\mathcal{H}$ on $K$ consecutive states, so we no longer find
a CTI from a chain of fewer than $K$ states inside $\mathcal{H}$. However, we can stay in the spurious loop for any $K$, so a larger window never removes its CTI unless the check also requires a loop-free path, at a cost that grows with $K$~\cite{sheeran2000checking}. We can only break such a loop with another helper, such as $h_1$ in
Fig. \ref{fig:trap}. In short:

\begin{figure*}[t]
  \centering
  \includegraphics[width=0.75\textwidth]{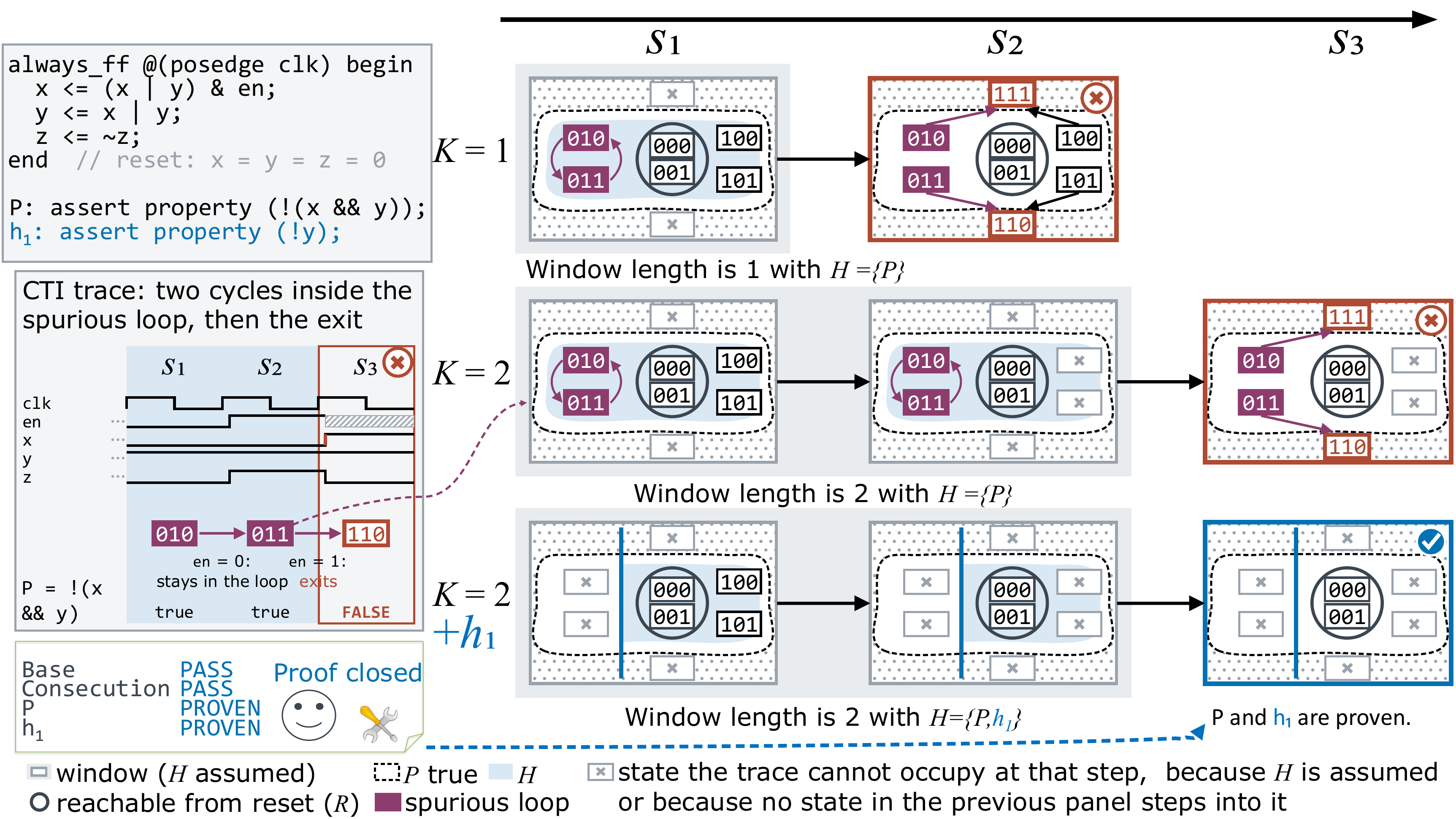}
  \caption{How a spurious loop and a chain differ as the window grows. We assume $\mathcal{H}$ for $K$
  cycles ($s_1,\dots,s_K$) and check $\mathcal{H}$ for the next cycle ($s_{K+1}$). The states 010 and 011 form a
  spurious loop, and the CTI chains start from 100 and 101. At $K=1$ both the loop and the chain give a CTI, and at $K=2$
  only the spurious loop remains. Widening the window removes the chain but not the loop, which we break by adding the helper $h_1$.
  }
  \label{fig:trap}
\end{figure*}

\begin{prop}
\label{prop:recurrent}
To establish a proof in \fra, every spurious loop needs to be broken by helper assertions, for any $K$.
\end{prop}

We therefore query for spurious loops directly with a bounded \emph{cover} query. As in the consecution check,
the engine starts from an arbitrary state and searches for a path $s_0,\dots,s_n$ with
\[
\underbrace{\textstyle\bigwedge_{i<n} T(s_i,s_{i+1})}_{\text{path}}\;\wedge\;\underbrace{s_\ell=s_0}_{\text{loop}}\;\wedge\;\underbrace{\textstyle\bigwedge_{i<n} P(s_i)\;\wedge\;\neg P(s_n)}_{\text{exit to }\neg P},\quad \ell<n,
\] 
where $T$ is the transition relation under the design's assumptions and the current helpers. The path goes around a loop of $\ell$ states on which $P$ holds and then exits to a state where $P$ fails. We pose the query as one SVA cover sequence, in which the state repeats with period $\ell$ and $P$ then holds for at most $K-1$ more cycles before it fails. The engine thus searches every path length $n$ up to $\ell+K$. We query $\ell\in\{1,2\}$ to capture common hardware patterns at low engine cost. Longer loops still reach the agent as raw CTIs. The query compares full states, every register.

We use spurious loops to guide the agent, and Algorithm \ref{alg:loops} contrasts this flow with the flow that uses raw CTIs. On top of the basic flow (Sec. \ref{sec:flow}), we also let the agent call the base check, the consecution check and the spurious-loop search as tools during the run. Once the search finds no loop, or times out, we let the agent choose the window $K$ for the remaining CTIs. \afra is \textbf{\textit{agnostic}} to the backend proof engine, and in this work we use \jasper{}~\cite{cadence2026jasper}, a commercial tool commonly used for sign-off.

\begin{algorithm}[t]
  \caption{\footnotesize Proof construction from raw counterexamples, as in CTI-guided LLM work~\cite{su2026cill}, and from spurious loops (\afra).
  $\mathit{Cons}_K(X)$: every assertion in $X$ passes the consecution check over $K$ cycles. We also check the
  helpers for the base case.}
  \label{alg:loops}
  \scriptsize
  \begin{minipage}[t]{0.45\columnwidth}
    \textbf{Counterexample-guided}
    \begin{algorithmic}[1]
      \State $H \gets \emptyset$
      \While{$\neg\mathit{Cons}_K(\{P\} \cup H)$}
        \State $C \gets$ CTIs of $\{P\} \cup H$
        \State $H \gets H \cup \mathit{Agent}(C, H)$
      \EndWhile
      \State \Return $H$
    \end{algorithmic}
  \end{minipage}\hfill
  \begin{minipage}[t]{0.53\columnwidth}
    \textbf{\afra}
    \begin{algorithmic}[1]
      \State $H \gets \emptyset$
      \While{\ours{a spurious loop $\lambda$ is found}}
        \Repeat\ $H \gets H \cup \mathit{Agent}(\ours{\lambda}, H)$
        \Until{\ours{$\mathit{Cons}_K(H)$} \Comment{\ours{without $P$}}}
      \EndWhile
      \Repeat\ $C \gets$ CTIs
        \State $H, \ours{K} \gets \mathit{Agent}(C, H)$ 
      \Until{$\mathit{Cons}_K(\{P\} \cup H)$}
      \State \Return $H$
    \end{algorithmic}
  \end{minipage}
\end{algorithm}

\subsection{Proof-Completion Training with a Progress Measure}
\label{sec:train}

To address sparse progress feedback, we train a customized model with a dense reward from proof progress on the graph (Fig. \ref{fig:anchor}), also to cut the cost of commercial agents.

For a given task, we define the progress measure $\Phi_{\text{task}}(H)$ of a partial
helper set $H$ as the work left to proof closure, which we use to construct a potential-based reward, where an $H$ that achieves proof closure has $\Phi_{\text{task}}(H)=0$, while $\Phi_{\text{task}}(\emptyset)=1$. We
consider three alternatives: (1) the bounded proof depth of the target under $H$; (2) the runtime to find a spurious
loop, which grows as fewer loops remain; and (3) the distance from $H$ to a golden proof $H^*$ on the dependency graph of $H^*$. Sec. \ref{sec:phi-eval} shows that (3) best 
distinguishes successful from unsuccessful runs,
so we use it as the dense reward. Specifically, we assume $H^*$, check each helper of $H$, and keep those that hold as $H'$. We then assume $H'$ and
the targets, check each helper of $H^*$, and count the golden helpers that still fail as the work left. We weigh
each block $i$ of these helpers by its $n_i$ open helpers and its layer $d_i$, and skip blocks with no open helper:
\begin{equation}
\begin{gathered}
D(H)=\sum_{i} g(n_i)\Big(1+\frac{d_i}{d_{\max}}\Big)+[\text{a target is open}],\\
\Phi_{\text{task}}(H)=\frac{\ln\big(1+D(H)\big)}{\ln\big(1+D(\emptyset)\big)},
\end{gathered}
\end{equation}
where $g(n)=\min(2^{n-1}, n+26)$, so the reward is not dominated by a single large block. The last term adds one unit of work while a target is unproven, so $\Phi_{\text{task}}(H)=0$ only at proof closure. We set the reward of a trajectory to $1-\Phi_{\text{task}}(H)$ minus a penalty for the share of its helpers that are false. Building the graph of $H^*$ takes $|H^*|$ deletions, but we do it offline once per golden proof. Each evaluation of $\Phi(H)$ then needs only $O(|H^*|+|H|)$ consecution queries.

We train Qwen3.8-27B with supervised fine-tuning (SFT) and then rejection sampling fine-tuning (RFT)~\cite{yuan2023scaling}. We use agents to build training tasks from designs outside the test set, composing modules, writing formal testbenches and targets across domains and scales, and every target goes through at least 4 hours of \jasper{} runs so that we work on hard targets. We measure the 5-gram Jaccard similarity between each training design and each test design, and ensure it is below 0.8. We also remove the top 2, 4 or 6 layers of golden proofs to obtain curriculum tasks, in which the agent rebuilds the removed helpers (Fig. \ref{fig:anchor}). We use the progress measure as a filter and keep the trajectories that reach a high reward.

%% file: sections/eval.tex
\section{Evaluation}
\label{sec:eval}

\subsection{Setup}
\label{sec:setup}

\begin{figure}[t]
  \centering
  \includegraphics[width=\columnwidth]{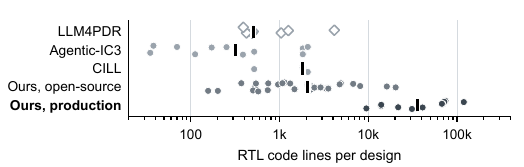}
  \caption{RTL code lines per design, evaluated in our work and in prior LLM proof-closure work, with LLM4PDR's sizes as its paper reports them.}
  \label{fig:scale}
\end{figure}

We define a task as a design, its formal testbench and its target properties. Human experts wrote the targets of the production tasks for sign-off and reviewed the targets of the open-source tasks to have sign-off quality. We note that we do not claim \afra achieves automated sign-off, which would require creating holistic assertions and is orthogonal to this work, and we do not consider bug hunting. The targets are indeed hard, since the HWMCC'25 winner rIC3 proves only 10.8\% of them. We evaluate on 587 targets in 50 tasks from 13 design families (Table \ref{tab:testset}), including eight open-source families spanning CPUs, GPUs and manycores, NICs and crossbars, and accelerators, and five anonymized production families (P1-P5) from industrial GPU and CPU designs. P1 is a configurable crossbar switch with scheduler, P2-P4 are GPU memory hubs, and P5 is a unit of a CPU coherent interconnect. Our production designs reach up to \textbf{118.9K} lines of RTL code, \textbf{29$\times$} the largest design in prior LLM proof-closure work~\cite{su2026cill,fan2026agentic,hu2026enhancing} (Fig. \ref{fig:scale}).

We evaluate \afra against four categories of State-of-the-Art (SoTA) approaches: (1) model checkers with no agent: \jasper{} and the HWMCC'25 winners Pono and rIC3~\cite{mann2021pono,su2025ric3,hwmcc2025}; (2) open coding agents: Qwen3.8-27B and Kimi~K3 (xhigh) in the Pi harness~\cite{zechner2025pi}; (3) closed coding agents: Codex with GPT-6.1-sol and GPT-5.6-sol (xhigh) and Claude Code with Opus~5.5 (max); and (4) customized agent flows: CILL~\cite{su2026cill}, LLM4PDR~\cite{hu2026enhancing}, Agentic-IC3~\cite{fan2026agentic} and ChatSVA~\cite{fu2026chatsva}, all run with GPT-6.1-sol (xhigh), as \afra{} is. We include ChatSVA, which targets assertion generation rather than proof closure, to show that helper strengthening needs a different capability. To run the model checkers and the customized flows on all 50 tasks, we translated each design to BTOR2 with Yosys~\cite{wolf2016yosys} and its yosys-slang frontend~\cite{yosysslang} for Pono, rIC3, Agentic-IC3 and LLM4PDR, and adapted CILL's frontend. Beyond its frontend, we run CILL as released, with its rIC3 and prompt loop unmodified. The one exception is P5, which none of these frontends can read. We rebuilt Agentic-IC3 and LLM4PDR from their papers, since neither released code. \afra's prompt describes only the goal and the workspace, with no task-specific skills. We further check every baseline's proofs with \jasper{}, and the comparison with GPT-6.1-sol in Codex keeps the model and the engine fixed.

We give every method 4 hours per task ($T=4$ hours), four times the 1-hour limit of HWMCC'25~\cite{hwmcc2025}, and report the percentage of targets the engine proves, averaged over families. The only exception is P5, which is our most complex task, and according to engineering practice, we give it a 24-hour budget. We set $T_1=900$ seconds and $T_2=2000$ seconds. The agent runs behind Table \ref{tab:main} alone took over 3000 agent-hours, and serving our trained 27B for its evaluation took about 1900 A100 GPU-hours, which limits the ablations we can afford. We use \jasper{} 2025.06 in its automatic engine mode, in which \jasper{} selects and orchestrates its proof engines itself, and run the engines in parallel with ProofGrid, on 4 cores for open-source tasks and 16 cores for production tasks, consistent throughout all methods, and on 100 cores for the \jasper{} baseline on production. We do not observe long license wait times.

\begin{table}[t]
  \centering
  \caption{\footnotesize Diversity and coverage of design and properties. KLoC: thousands of RTL lines. The targets cover ordering, data integrity, coherence, deadlock freedom and fairness, etc. All targets are safety properties, so deadlock freedom and fairness take the form of bounded waits.}
  \label{tab:testset}
  \input{tables/table1_test_set.tex}
\end{table}

\subsection{Comparing with SoTA}
\label{sec:main}

\input{tables/table2_main.tex}

On the 50 tasks, \afra with GPT-6.1-sol proves 80.5\% of the targets on average over the 13 families, more than twice the 36.4\% of the same model in Codex (Table \ref{tab:main}). The strongest customized flow, CILL, proves 25.3\%, and the HWMCC'25 winners prove at most 10.8\%. On the production families, only \afra proves any target on P3 and P4. For P5, only \afra{} proves all five targets. The gain is also large on open-source families. On Corundum, NVDLA and OpenTitan, the targets proven increase from 52.1\% to 93.6\%, from 0\% to 87.5\% and from 27.8\% to 100\%.

\begin{figure}[t]
  \centering
  \includegraphics[width=\columnwidth]{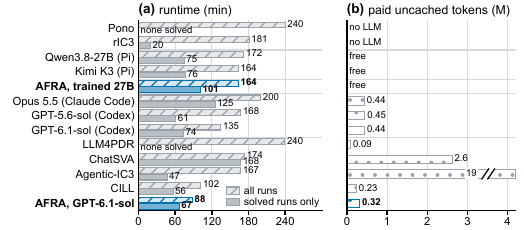}
  \caption{Runtime (agent + proof check) and paid uncached tokens per task, geometric means over the runs of Table \ref{tab:main}. (a) Runtime
  over all runs and over the solved runs only, which reach proof closure. (b) Paid uncached tokens, the input tokens not read from the cache plus the output tokens. We mark the open models as free since they pay no API fee.}
  \label{fig:runtime}
\end{figure}

\begin{figure}[t]
  \centering
  \includegraphics[width=\columnwidth]{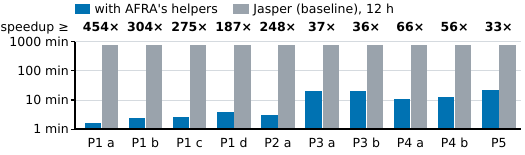}
  \caption{Proof checking time on each production task \afra fully proven, with \afra's helpers and with \jasper{} alone. \jasper{} alone finds no full proof of these tasks in 12 hours, so the speedups are lower bounds. \afra's proofs use on average 103, 121, 121, 84 and 994 helpers per task on P1 to P5.}
  \label{fig:prooftime}
\end{figure}

We report the runtime and paid uncached tokens per task in Fig.~\ref{fig:runtime}. Over all runs, \afra with GPT-6.1-sol takes 88 minutes per task, compared with 102 minutes for CILL, the strongest customized flow. For solved tasks, \afra takes 67 minutes, a reasonable solve time from an engineering perspective. With a slight token overhead, \afra proves $80.5\%$ of the targets, over $3\times$ CILL's $25.3\%$.

We also measure how long \jasper{} takes to check the finished proofs of the 10 production tasks that \afra fully proves (Fig. \ref{fig:prooftime}). With \afra's helpers, \jasper{} proves every target of each task in 1.6 to 21 minutes. With \jasper{} alone, we find no proof of any of these tasks in 12 hours, and even with 30 hours, \jasper{} shows no progress. \afra's helpers speed up the proofs by at least $33\times$ and up to $454\times$. This speedup is for checking the targets with helpers, while the end-to-end runtime is in Fig. \ref{fig:runtime}. A fast check is also valuable in production, where it is rerun after every design change.

\subsection{Progress Measure and the Trained Model}
\label{sec:phi-eval}

\begin{figure}[t]
  \centering
  \includegraphics[width=\columnwidth]{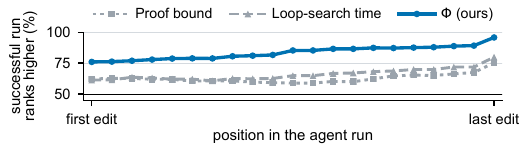}
  \caption{Three progress measures on agent runs across our training tasks. At each position in a run, we plot the accuracy of ranking a run that reaches proof closure above a run that does not, with a tie counted as half.}
  \label{fig:phi}
\end{figure}

A reward for training must rise as the agent makes progress, and it must not call a failed run done. We compare $\Phi$ with the two alternatives of Sec. \ref{sec:train}, the bounded proof depth and the time to find a spurious loop, on the same agent runs across our training tasks (Fig. \ref{fig:phi}). At each position in a run, we pair every run that fully proves its targets with every run that never does, and count how often the successful run ranks higher. With $\Phi$, the successful run ranks higher in 76\% of the pairs after the agent's first edit, versus 62\% with the bounded depth and 61\% with the loop-search time. Both alternatives need a limit to stay between 0 and 1. The bounded depth has no natural end, so we set an upper bound. For the loop search, it slows as fewer loops remain, so we set a timeout. Once the limit is reached, the measure reads complete although the targets remain unproven. $\Phi$ needs no timeout or upper bound, and we use $\Phi$ as the reward.

\begin{figure}[t]
  \centering
  \includegraphics[width=\columnwidth]{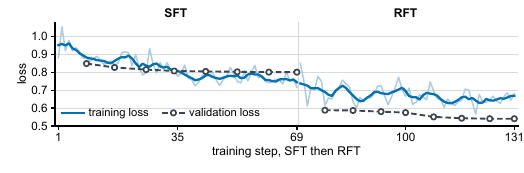}
  \caption{Training and validation loss of the trained 27B, first in SFT and then in RFT. The light line is the per-step training loss, and the solid line is its running mean over 5 steps.}
  \label{fig:loss}
\end{figure}

We fine-tune all parameters of Qwen3.8-27B on 16 nodes of 8 A100 GPUs, with sequences of up to 262K tokens, for 2 epochs of SFT at a peak learning rate of $10^{-5}$ and 2 of RFT at $5\times10^{-6}$, and serve it with vLLM~\cite{kwon2023vllm} at 8-way tensor parallelism. We implement the training with NeMo-RL~\cite{nemorl} on its Megatron backend~\cite{shoeybi2019megatron}. We train on full multi-turn agent trajectories with every tool call and its result, and compute the loss only on the agent's output tokens. When a trajectory fills the context, the harness compacts it, and we train on the compacted context as is. For SFT, we collect 950 Kimi K3 trajectories and keep 391 that pass the filter of Sec. \ref{sec:train}. We train on 346 of them from 124 tasks and validate on the other 45. For RFT, we collect 440 trajectories of our SFT model and keep 104, and to avoid forgetting~\cite{scialom2022fine}, we mix in 190 Kimi K3 trajectories, 294 in total from 109 tasks. We balance the data over tasks. SFT caps each task at 40 trajectories, and RFT picks part of the Kimi K3 trajectories round-robin over tasks. Over the two epochs, we train on 167.1M tokens in SFT and 147.6M in RFT. We show the training and validation loss of SFT and RFT in Fig. \ref{fig:loss}.

With the trained 27B, \afra proves 32.7\% of the targets, against 22.8\% for the untrained 27B in the same harness and 12.8\% in the Pi harness. It outperforms GPT-5.6-sol (20.4\%) and Opus~5.5 (21.1\%) in their coding agents and its teacher Kimi K3 (25.4\%) used in SFT (Table \ref{tab:main}). In the RFT phase, we can train on 5.2$\times$ as many tokens with filtering by the progress reward, compared with filtering the trajectories that fully prove their task. Served on our own nodes, the 27B costs no API fees and takes 164 minutes per task, matching Kimi~K3 in Pi (Fig. \ref{fig:runtime}), and newer GPUs will be much faster.

\subsection{Harness Ablation}
\label{sec:ablation}

In Table \ref{tab:ablation}, we add \afra's harness contributions one at a time to a naive harness, with GPT-6.1-sol in every row. With the modeling code and then the spurious-loop search, we raise the production mean from 14.0\% to 36.7\% and then to 74.2\%, and we prove targets on P4 and P5 only with the spurious-loop search.

\begin{table}[t]
  \centering
  \caption{Target properties proven (\%) as we add \afra's harness contributions, with GPT-6.1-sol and two runs. Mean: over P1-P5. 
  }
  \label{tab:ablation}
  \input{tables/table3_ablation.tex}
\end{table}

\begin{figure}[t]
  \centering
  \includegraphics[width=\columnwidth]{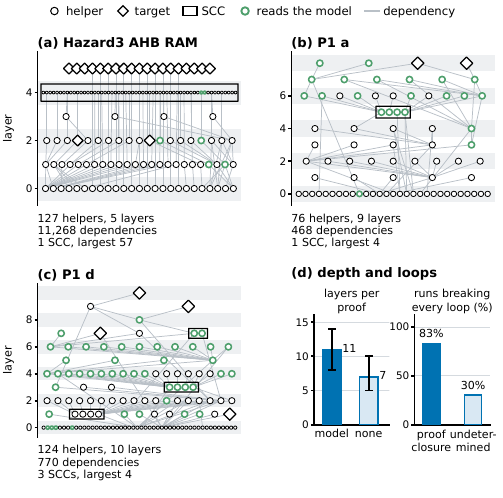}
  \caption{(a)-(c) Dependency graphs of three golden proofs by layer, with generate-loop copies counted once and lines for the transitive reduction of the dependencies. (d) Median layers of golden proofs with and without modeling code, with quartiles, and the share of runs of \afra with GPT-6.1-sol that break every spurious loop the search finds.}
  \label{fig:graphs}
\end{figure}

\subsection{Case Study}
\label{sec:case}

We study P1 d, the largest P1 task, where \afra proves all 3 hard targets in both runs. In its dependency graph (Fig. \ref{fig:graphs}(c)), the target that \jasper{} alone proves is at layer 1, while the 3 hard targets are at layers 7, 9 and 10. Proofs this deep usually come with modeling code. Over our golden proofs, those with modeling code have a median of 11 layers, against 7 without it (Fig. \ref{fig:graphs}(d)), and on both P1 tasks, the helpers that read the behavioral model are in the upper layers, as they need to relate signals from multiple modules. On a training task built around the Hazard3 core~\cite{wrenhazard3} (Fig. \ref{fig:graphs}(a)), 57 helpers hold only together, so one missing helper makes all 57 fail and buries the root cause in their CTIs.

The spurious-loop search also changes what the agent receives. On the tasks that \afra fully proves and CILL does not, the agent receives about 50K tokens of loops per run in its feedback files, instead of about 603K tokens of raw counterexamples, 98\% of which have no repeated state, so the agent cannot tell a loop from a chain. Minimizing a counterexample, as CILL does, shortens it but still shows no repeated state, yet a loop has to be blocked by a helper (Proposition \ref{prop:recurrent}).

With GPT-6.1-sol, the agent writes modeling code more often on the production families. Among the runs with proof closure, 83\% break every spurious loop the search finds, against 30\% of the undetermined runs (Fig. \ref{fig:graphs}(d)). A check refutes 0.9\% of the proposed helpers from reset, and such false helpers never lead to proof closure (Sec. \ref{sec:flow}).

%% file: tables/table1_test_set.tex
\setlength{\tabcolsep}{4pt}
\begin{tabular}{@{}llrrr@{}}
\toprule
Domain & Families & Tasks & Targets & KLoC \\
\midrule
CPU & SERV, CVA6 & 15 & 79 & 2.1-20.3 \\
GPU, manycore & Nyuzi, OpenPiton & 11 & 102 & 0.2-16.2 \\
NIC, crossbar & Corundum, PULP & 7 & 50 & 0.5-2.3 \\
Accel., SoC & NVDLA, OpenTitan & 4 & 13 & 1.1-4.8 \\
Production & P1-P5 & 13 & 343 & 9.5-118.9 \\
\midrule
Total & 13 families & 50 & 587 & 0.2-118.9 \\
\bottomrule
\end{tabular}

%% file: tables/table2_main.tex
\begin{table*}[t]
\centering
\caption{Percentage of target properties proven in each family
(mean $\pm$ std over two runs). Mean: over the 13 families, counting N/A cells as 0. UD: undetermined. N/A: cannot elaborate this design. Bold: best closed-model entry. Underline: best open-model entry.}
\label{tab:main}
\setlength{\tabcolsep}{2pt}
\resizebox{\textwidth}{!}{
\begin{tabular}{@{}lcccccccccccccc@{}}
\toprule
 & \multicolumn{8}{c}{Open-source Design Family} & \multicolumn{5}{c}{Production Design Family} & \multicolumn{1}{c}{Overall} \\
\cmidrule(lr){2-9}\cmidrule(lr){10-14}\cmidrule(l){15-15}
Design family & SERV & CVA6 & Corundum & OpenPiton & Nyuzi & NVDLA & OpenTitan & PULP & P1 & P2 & P3 & P4 & P5 & Mean \\
\# tasks & 8 & 7 & 6 & 7 & 4 & 2 & 2 & 1 & 4 & 2 & 3 & 3 & 1 & 50 \\
\# target properties & 16 & 63 & 47 & 54 & 48 & 4 & 9 & 3 & 10 & 4 & 36 & 288 & 5 & 587 \\
\midrule
\multicolumn{15}{@{}l}{\textit{Model checkers, no agent}} \\
Pono & UD & \mstd{1.6}{0.0} & UD & UD & UD & UD & UD & UD & \mstd{10.0}{0.0} & UD & UD & UD & N/A & \mstd{0.9}{0.0} \\
Jasper & \mstd{18.8}{0.0} & \mstd{11.1}{0.0} & UD & \mstd{1.9}{0.0} & UD & UD & UD & UD & \mstd{10.0}{0.0} & UD & UD & UD & UD & \mstd{3.2}{0.0} \\
rIC3 & \mstd{50.0}{0.0} & \mstd{30.2}{1.6} & \mstd{17.0}{0.0} & \mstd{20.4}{0.0} & \mstd{12.5}{0.0} & UD & UD & UD & \mstd{10.0}{0.0} & UD & UD & UD & N/A & \mstd{10.8}{0.1} \\
\midrule[\heavyrulewidth]
\multicolumn{15}{@{}l}{\textit{Open coding agents, highest reasoning effort}} \\
Qwen3.8-27B (Pi) & \mstd{62.5}{6.2} & \mstd{35.7}{0.8} & \mstd{3.2}{1.1} & \mstd{20.4}{5.6} & \mstd{24.0}{9.4} & UD & \mstd{11.1}{0.0} & UD & \mstd{10.0}{0.0} & UD & UD & UD & UD & \mstd{12.8}{1.6} \\
Kimi K3 (Pi) & \mstd{\underline{87.5}}{0.0} & \mstd{\underline{48.4}}{4.0} & \mstd{\underline{23.4}}{2.1} & \mstd{25.9}{0.0} & \mstd{51.0}{1.0} & UD & UD & \mstd{\underline{83.3}}{16.7} & \mstd{10.0}{0.0} & UD & UD & UD & UD & \mstd{25.4}{0.7} \\
\cmidrule(l){2-15}
\multicolumn{15}{@{}l}{\textit{Customized agent flow, open model}} \\
AFRA, Qwen3.8-27B (Pi) & \mstd{12.5}{6.2} & \mstd{31.7}{6.3} & \mstd{2.1}{0.0} & \mstd{25.9}{14.8} & \mstd{\underline{56.2}}{0.0} & \mstd{\underline{75.0}}{0.0} & \mstd{11.1}{0.0} & \mstd{66.7}{0.0} & \mstd{15.0}{5.0} & UD & UD & UD & UD & \mstd{22.8}{0.6} \\
\textbf{AFRA, trained 27B (Pi)} & \mstd{75.0}{6.2} & \mstd{41.3}{1.6} & \mstd{\underline{23.4}}{8.5} & \mstd{\underline{50.0}}{0.0} & \mstd{51.0}{7.3} & \mstd{37.5}{37.5} & \mstd{\underline{50.0}}{27.8} & \mstd{66.7}{0.0} & \mstd{\underline{30.0}}{0.0} & UD & UD & UD & UD & \mstd{\underline{32.7}}{0.8} \\
\midrule[\heavyrulewidth]
\multicolumn{15}{@{}l}{\textit{Closed coding agents, highest reasoning effort}} \\
Opus 5.5 (Claude Code) & \mstd{62.5}{12.5} & \mstd{41.3}{1.6} & \mstd{14.9}{8.5} & \mstd{25.9}{1.9} & \mstd{41.7}{14.6} & UD & \mstd{27.8}{5.6} & \mstd{50.0}{50.0} & \mstd{10.0}{0.0} & UD & UD & UD & UD & \mstd{21.1}{5.7} \\
GPT-5.6-sol (Codex) & \mstd{71.9}{3.1} & \mstd{49.2}{11.1} & \mstd{20.2}{3.2} & \mstd{25.9}{0.0} & \mstd{43.8}{4.2} & UD & \mstd{11.1}{0.0} & \mstd{33.3}{33.3} & \mstd{10.0}{0.0} & UD & UD & UD & UD & \mstd{20.4}{0.9} \\
GPT-6.1-sol (Codex) & \mstd{\mathbf{100.0}}{0.0} & \mstd{53.2}{2.4} & \mstd{52.1}{7.4} & \mstd{25.9}{0.0} & \mstd{60.4}{4.2} & UD & \mstd{27.8}{5.6} & \mstd{83.3}{16.7} & \mstd{70.0}{0.0} & UD & UD & UD & UD & \mstd{36.4}{0.8} \\
\cmidrule(l){2-15}
\multicolumn{15}{@{}l}{\textit{Customized agent flows, GPT-6.1-sol, highest reasoning effort}} \\
LLM4PDR (Codex) & UD & UD & UD & \mstd{3.7}{0.0} & UD & UD & UD & UD & UD & UD & UD & UD & N/A & \mstd{0.3}{0.0} \\
ChatSVA (Codex) & \mstd{25.0}{6.2} & \mstd{8.7}{2.4} & \mstd{1.1}{1.1} & \mstd{2.8}{0.9} & UD & UD & UD & UD & \mstd{10.0}{0.0} & UD & UD & UD & UD & \mstd{3.7}{0.1} \\
Agentic-IC3 (Codex) & \mstd{25.0}{0.0} & \mstd{15.1}{11.9} & \mstd{10.6}{4.3} & \mstd{20.4}{0.0} & \mstd{12.5}{0.0} & UD & \mstd{33.3}{33.3} & \mstd{66.7}{0.0} & UD & UD & UD & UD & N/A & \mstd{14.1}{1.3} \\
CILL (Codex) & \mstd{90.6}{3.1} & \mstd{23.8}{0.0} & \mstd{20.2}{5.3} & \mstd{18.5}{0.0} & \mstd{47.9}{0.0} & \mstd{25.0}{25.0} & \mstd{27.8}{5.6} & UD & \mstd{50.0}{10.0} & \mstd{25.0}{25.0} & UD & UD & N/A & \mstd{25.3}{2.8} \\
\textbf{AFRA, GPT-6.1-sol (Codex)} & \mstd{\mathbf{100.0}}{0.0} & \mstd{\mathbf{57.1}}{0.0} & \mstd{\mathbf{93.6}}{6.4} & \mstd{\mathbf{37.0}}{14.8} & \mstd{\mathbf{100.0}}{0.0} & \mstd{\mathbf{87.5}}{12.5} & \mstd{\mathbf{100.0}}{0.0} & \mstd{\mathbf{100.0}}{0.0} & \mstd{\mathbf{100.0}}{0.0} & \mstd{\mathbf{50.0}}{0.0} & \mstd{\mathbf{66.7}}{0.0} & \mstd{\mathbf{54.2}}{18.1} & \mstd{100.0}{0.0} & \mstd{\mathbf{80.5}}{1.7} \\
\bottomrule
\end{tabular}}
\end{table*}

%% file: tables/table3_ablation.tex
\setlength{\tabcolsep}{2pt}
\scriptsize
\resizebox{\columnwidth}{!}{
\begin{tabular}{@{}lcccccc@{}}
\toprule
Harness & P1 & P2 & P3 & P4 & P5 & \makecell{Mean\\(on Prod.)} \\
\midrule
Naive harness & \mstd{70.0}{0.0} & UD & UD & UD & UD & \mstd{14.0}{0.0} \\
+ behavioral model & \mstd{\mathbf{100.0}}{0.0} & \mstd{\mathbf{50.0}}{0.0} & \mstd{33.3}{33.3} & UD & UD & \mstd{36.7}{6.7} \\
+ spurious loop & \mstd{\mathbf{100.0}}{0.0} & \mstd{\mathbf{50.0}}{0.0} & \mstd{\mathbf{66.7}}{0.0} & \mstd{\mathbf{54.2}}{18.1} & \mstd{\mathbf{100.0}}{0.0} & \mstd{\mathbf{74.2}}{3.6} \\
\bottomrule
\end{tabular}}

%% file: sections/related.tex
\section{Related Work}
\label{sec:related}
We identify five major categories of related work. (i) Assertion generation with LLMs from
specifications~\cite{autosva2,wu2026assertllm2,assertionforge,pulavarthi2025assertionbench,kang2025fveval,fu2026chatsva} and debugging~\cite{bai2025fvdebug,bai2026rtlinsight}. (ii) In the category of proof construction via helper-strengthening, helpers are written manually~\cite{cadence2026jasper,hernandez2025,gilday2021human,mcmillan1998verification}, or inferred by code~\cite{kamath2024leveraging} or generated with LLMs~\cite{peled2025large,su2026cill}. \afra belongs to this category. (iii) In the LLM-in-the-engine category, an LLM is used to guide the PDR or IC3 search~\cite{hu2026enhancing,fan2026agentic}, to tune IC3's heuristics~\cite{miao2026ic3}, or to abstract the RTL inside a counterexample-guided abstraction refinement loop~\cite{yan2026neuroabs}. (iv) In the proof-engine category, engines range from k-induction to IC3 and PDR~\cite{bradley2011sat,een2011efficient}, with rIC3 and Pono as the recent SoTA~\cite{su2025ric3,mann2021pono}. Classical engines handle CTIs internally, as k-induction rules out repeated states along a path~\cite{sheeran2000checking} and IC3 learns clauses that block each one~\cite{bradley2011sat,gurfinkel2017k}. Specialized approaches for specific checkers~\cite{biere2002liveness,xmasdeadlock} also belong to this category. (v) In the SAT-solver category, the solver can be tuned for IC3~\cite{su2025gipsat}, as in rIC3. Dependency graphs of invariants are not new~\cite{ivrii2014small}, but we are the first to use them to study where agents fail. Lean-based approaches~\cite{lyu2026rtl2lean,yang2026circuitprover} can express all RTL code but also require agent-dependent proof construction with limited progress indication. Prior work trains models for related hardware and software verification tasks~\cite{wu2026qimeng,miao2026legend,yan2025re,zhang2026llm4cov}. 

Unlike the closest related work in category (ii), \afra uses spurious-loop search and modeling code to construct much more complex proofs, and trains a much smaller model that performs competitively on this task. To our knowledge, we are the first to train a custom LLM agent for unbounded RTL proof closure.

%% file: sections/disclosure.tex
\section{Content Generated by AI}

We use Claude Opus 5.5 to polish text and to help with parts of the implementation, all of which the authors reviewed.

%% file: sections/conc.tex
\section{Acknowledgement}

The authors would like to thank Hima Swetha, Chirag Agarwal, Ping Yeung, Prosenjit Chatterjee, and Vigyan Singhal for their support throughout the project, Leo (Ang) Li for valuable discussions, and Igor Markov and Jose Renau for their guidance and for reviewing this work. They also thank Ayush Vajpeyi and Xiaoyu Yan for promoting FRA in engineering practice, which helped motivate this work.

\section{Conclusion}
\label{sec:conc}

We present \afra, an agentic proof-construction framework that turns bounded proofs into unbounded proofs.
By viewing a full proof as a proof dependency graph, we find that agents struggle to build deep and densely connected proofs and receive sparse progress feedback. \afra builds deep and densely connected proofs with behavioral modeling and spurious-loop search, and trains a model with a dense reward that measures proof progress on the graph. On 587 targets, \afra{} achieves SoTA proof closure results over strong baselines. 
Beyond the harness improvements, we also show the early promise of a specialized model. 
Future work includes scaling up the agent, improving the custom model, and extending \afra to bug discovery and RTL repair during proof construction, moving toward automated sign-off.

%% file: sections/appendix.tex
\appendices

\section{Implementation Details}
\label{app:impl}

We discuss several implementation details of the AFRA harness.

\textbf{Agent sessions.} We run every agent in headless mode, Codex as \texttt{codex exec}, Claude Code as
\texttt{claude -p} and Pi in its JSON mode, inside a sandbox, so that each agent only has access to its own workspace.
When a session stops early, for example on a provider error, we resume the same session with a prompt that points the
agent to its journal and the time left.

\textbf{Context and budgets.} We run GPT-6.1-sol with no limit on its number of tool calls and with its default context length and compaction settings. The 27B models have a shorter context length, so we handle them differently. We cut the work of the 27B models into rounds, so that we can measure the progress reward after each round. We further limit each round to at most 60 tool calls, 80 minutes of rollout time and one context compaction. After each round, the harness checks the helper file with \jasper{}, writes the CTIs and reports to the agent's workspace, and the next round starts from the updated workspace.

\textbf{Workspace.} Fig. \ref{fig:workspace} shows the workspace of the \afra agent.

\begin{figure}[H]
  \centering
  \fbox{\parbox{\dimexpr\columnwidth-2\fboxsep-2\fboxrule\relax}{\footnotesize
  \begin{tabular}{@{}l@{\hspace{0.8em}}l@{}}
    \texttt{BRIEF.md} & prompt \\
    \texttt{journal.md} & agent's log \\
    \texttt{pack/} & prepared by harness \\
    \quad\texttt{targets.tsv} & targets and their per-cycle conditions \\
    \quad\texttt{first\_rel.txt} & first proof check of targets \\
    \quad\texttt{first\_loop.txt} & first spurious loop \\
    \texttt{task/} & read-only copy of task package \\
    \quad\texttt{rtl/}, \texttt{sva/}, \texttt{lists/} & design, testbench, Tcl scripts \\
    \texttt{tools/prove.py} & tools \texttt{rel}, \texttt{loop}, \texttt{gate}, \texttt{cover}, \ldots \\
    \texttt{work/} & agent's helper sets, monitors, notes \\
    \texttt{.requests/} & tool requests and responses \\
  \end{tabular}}}
  \caption{Workspace of an \afra agent.}
  \label{fig:workspace}
\end{figure}

\textbf{Proof checks during agent run.} We implement the checks during the run as tools for the agent, mainly \texttt{rel} and \texttt{loop}, which the agent calls in its workspace, with a $T_1$ timeout. With \texttt{rel}, the agent checks the base case and the consecution of a set of properties, and receives TSV files of CTIs. With \texttt{loop}, the agent receives a spurious loop found by Jasper. We also allow the agent to do lookahead search with \texttt{loop -with}, so it can use helpers to block the current loop and the tool will search for a different one. With \texttt{cover}, the agent can ask whether a condition is reachable from reset. We also equip the agent with CTI lookup tools and tools that output the driver of a value.

\textbf{Spurious-loop query.} We search for a spurious loop of period $\ell$ with the cover property of Algorithm \ref{alg:cover}, with the input assumptions and the helpers of $\mathcal{H}$ assumed on every cycle. We generate the query from the list of registers that \jasper{} reports for the elaborated design, which includes the agent's modeling code. In the cover query, \texttt{s == \$past(s, $\ell$)} is the conjunction of \texttt{r == \$past(r, $\ell$)} over every register \texttt{r} of the design.

\begin{algorithm}[H]
  \caption{\footnotesize Cover query for a spurious loop of period $\ell$ and window $K$. $P$: the conjunction of the targets' per-cycle conditions. $s$: every register.}
  \label{alg:cover}
\begin{lstlisting}[style=sva]
cover property (
  P
  ##1 (P && s != $past(s))       // for @$\ell = 2$@
  ##1 (P && s == $past(s, @$\ell$@))
  ##1 P [*0:K-1] ##1 !P
);
\end{lstlisting}
\end{algorithm}

A cover trace traverses the loop before exiting to a state that violates $P$. Lines 2-4 describe the traversal. Line 3 is included for $\ell=2$, and its state change excludes loops of period 1. Line 4 returns to the state of $\ell$ cycles earlier, which closes the loop, so the loop can be sustained indefinitely by repeating the same input sequence. Afterward, $P$ holds for zero to $K-1$ additional cycles before failing (line 5). As in the consecution check, the search starts from an arbitrary state. If $P$ is an invariant, every state on a cover trace is therefore unreachable from reset. For example, for the design in Fig.~\ref{fig:trap} with $\mathcal{H}=\{P\}$, $\ell=2$, and $K=1$, the trace $011, 010, 011, 110$ satisfies the query, with $\mathit{en}=0$ for the first two transitions and $\mathit{en}=1$ for the last. The loop states are unreachable because $y$ remains 0 from reset. Adding the helper $h_1$ excludes this loop.

\textbf{Targets in the cover query.} The cover query checks $P$ cycle by cycle, but a target property can span several cycles. To address this issue, we rewrite each target as a per-cycle condition, a Boolean expression for a single cycle, with the rules below:
\begin{center}\scriptsize
\begin{tabular}{@{}l@{\hspace{0.8em}}c@{\hspace{0.8em}}l@{}}
\lstinline[style=sva]{a |-> b} & $\rightarrow$ & \lstinline[style=sva]{!a || b} \\
\lstinline[style=sva]{a |=> b} & $\rightarrow$ & \lstinline[style=sva]{!\$past(a) || b} \\
\lstinline[style=sva]{disable iff (d) (a |=> b)} & $\rightarrow$ & \lstinline[style=sva]{d || $past(d) || !$past(a) || b} \\
\end{tabular}
\end{center}
For any other temporal operator that spans multiple cycles, such as a bounded wait, we let the agent write a monitor for each such property, a module bound into the design that records the history the property needs, and define the per-cycle condition over the monitor's signals. We then check that this per-cycle condition and the property are equivalent in both directions, from an arbitrary state including the monitor's extra state, so the equivalence is proven on the full state space.

\textbf{Proof dependency graph.} We list the pseudo code of constructing the proof dependency graph given golden proof in Algorithm \ref{alg:graph}.

\textbf{Anti cheating.} We bind the agent's SVA file into a fresh copy of the task and elaborate the design with it. Once per task, we record the list of the task's input assumptions. To ensure no unproven helpers can be assumed, before each check, we list the assumptions again and reject the check when the assumption list doesn't match.

\textbf{Agent prompts.} We show the structure of the agent's prompt in Fig. \ref{fig:prompt-structure}, and part of the prompts on spurious loops in Fig. \ref{fig:prompt-loop} and on behavioral modeling in Fig. \ref{fig:prompt-model}.

\begin{figure}[H]
  \centering
  \fbox{\parbox{\dimexpr\columnwidth-2\fboxsep-2\fboxrule\relax}{\footnotesize\raggedright
  \# Prove the shipped targets of \texttt{\{task\}}, with the prover in your hands\\[1pt]
  [\ldots]\\[1pt]
  - BASE: each property holds for the first K cycles from real reset;\\[1pt]
  - STEP: from any state where every property of the set held for K cycles, each property holds at the next cycle.\\[3pt]
  \#\# The targets\\[1pt]
  [one row per target]\\[3pt]
  \#\# Done\\[1pt]
  One \texttt{rel -joint} request whose set holds every target body [\ldots] and which answers PASS for every
  member. [\ldots]\\[3pt]
  \#\# Your workspace\\[1pt]
  [\texttt{pack/}, \texttt{task/}, \texttt{work/}, \texttt{journal.md}]\\[3pt]
  \#\# Your tools: \texttt{python3 tools/prove.py} [\ldots]\\[1pt]
  [\texttt{loop}, \texttt{gate}, \texttt{rel}, \texttt{values}, \texttt{why}, \texttt{cover}, \texttt{target},
  \texttt{status}]\\[1pt]
  K is your choice, [\ldots] Loops are removed by properties, never by K. [\ldots]\\[3pt]
  \#\# How to work\\[1pt]
  [steps 0 to 6, partly in Fig. \ref{fig:prompt-loop}]\\[3pt]
  \#\# SVA rules\\[1pt]
  Write implications as \texttt{!a || b}. No casts, no function calls [\ldots]\\[3pt]
  \#\# Rules\\[1pt]
  - Write only inside this folder [\ldots]\\[1pt]
  - Never change the package. No network.\\[1pt]
  - You have \texttt{\{hours\}} hours of wall time, tool waits included. [\ldots]}}
  \caption{Structure of the agent's prompt.}
  \label{fig:prompt-structure}
\end{figure}

\begin{figure}[H]
  \centering
  \fbox{\parbox{\dimexpr\columnwidth-2\fboxsep-2\fboxrule\relax}{\footnotesize\raggedright 
  2. Loops first. Read the current loop (\texttt{pack/first\_loop.txt}, or \texttt{loop}). Explain the stall as a chain of waits: who waits for whom, holding what, using the RTL. Then decide why this state cannot happen from reset: which count, credit, occupancy, ordering or ownership law does it break?\\[3pt]
  3. Write that law and the supports it needs to \texttt{work/setNN.tsv}, the target bodies first. Screen with \texttt{gate}, \texttt{values} or \texttt{cover}, then prove the set with \texttt{rel -joint -file}.\\[3pt]
  [\ldots]\\[3pt]
  5. When a set passes, run \texttt{loop -with} with all your passing properties to see the next loop. Repeat from step 2 until NO LOOP LEFT, then choose K and run the Done request.}}
  \caption{Agent's prompt on spurious loops.}
  \label{fig:prompt-loop}
\end{figure}

\begin{figure}[H]
  \centering
  \fbox{\parbox{\dimexpr\columnwidth-2\fboxsep-2\fboxrule\relax}{\footnotesize\raggedright 
  Some claims need somewhere that remembers: ``what comes out is what went in'' needs a record of what went in. [\ldots] Many good helpers rest on the design's own registers instead, or the checker's---a counter that stays in range, two flags that must agree, a one-hot that is one. Those declare nothing, and they are cheap.\\[3pt]
  But some claims need history the design does not keep. [\ldots] When the claim you need is of that kind, add the register-and then give the register its own invariant as a member of the set. [\ldots]\\[3pt]
  Model the question, not the machine. When you add state you are choosing what to remember, and copying the design's own structure is the expensive choice. [\ldots] A counter, an index, or a map replaces a traversal with arithmetic. Record a position at the moment it is assigned rather than deriving it by walking. Keep a count rather than re-deriving how much is there.}}
  \caption{Agent's prompt on behavioral modeling.}
  \label{fig:prompt-model}
\end{figure}

\section{Helper Case Study}
\label{app:categories}

We study the helpers in the proof closure \afra achieved, and identify the correlation between a helper's semantics and its position in the proof dependency graph.

We classify the helpers into eight categories with the help of an agent, which labels each group of helpers that differ only in their numbers after we parse them with slang. (1) \emph{Legal values}: a signal only takes allowed values, such as one-hot or within a range. (2) \emph{Control logic}: two control states never happen together, or one implies the other. (3) \emph{Pipelines and handshakes}: a stage holds what the previous stage held, or valid data stays stable until it is accepted. (4) \emph{FIFO pointers and counts}: a queue's pointers, count and flags agree. (5) \emph{Credit counts}: several counters add up. (6) \emph{Unique IDs}: no two live entries share an ID, tag, address or slot. (7) \emph{Model matches design}: the agent's model or the checker holds the same value as the design. (8) \emph{Data and order}: a tracked item's data and position match the design's storage. We put the helpers that fit none of these categories, such as a single flag of the agent's checker, into \emph{Other}. We show example helper assertions of each category in Table \ref{tab:examples}.

\begin{table}[t]
  \caption{Example helper assertions in each category.}
  \label{tab:examples}
\begin{lstlisting}[style=sva, numbers=none, frame=single, framesep=3pt, framerule=0.4pt,
  rulecolor=\color{black}, xleftmargin=3.4pt, xrightmargin=3.4pt]
// Legal values
$onehot0(buf_q[e].hit_oh)
available_credit[0] <= 3'd4
// Control logic
!ibus_cyc || !cnt_en
w_gnt |-> w_req
// Pipelines and handshakes
resp_valid && !resp_ready |=> resp_valid && $stable({resp_queue, ...})
retired |-> $past(last_bit && pc_running)
// FIFO pointers and counts
tail == head + count[2:0]
is_two_or_more_f == (count_f >= 3'd3)
// Credit counts
n_all == n_wf + n_fg + n_gr
accepted0 == emitted0 + {8'd0, count0}
// Unique IDs
!(live[1] && live[2]) || (ldbuf_q[1].trans_id != ldbuf_q[2].trans_id)
$onehot0(tid_owners[tid])
// Model matches design
chk.o_iptr == scoreboard.issue_pointer_q
chk.occ0 == {1'b0, fifo_count[0]}
// Data and order
accepted_xor == (inventory_xor ^ delivered_xor)
!(sym == 2'd0 && trk_v) || trk_data == fifo.data[rrp + trk_ahead[2:0]]
// Other
testbench.u_inv.ok_stream
\end{lstlisting}
\end{table}

\begin{figure*}[t]
  \centering
  \includegraphics[width=\textwidth]{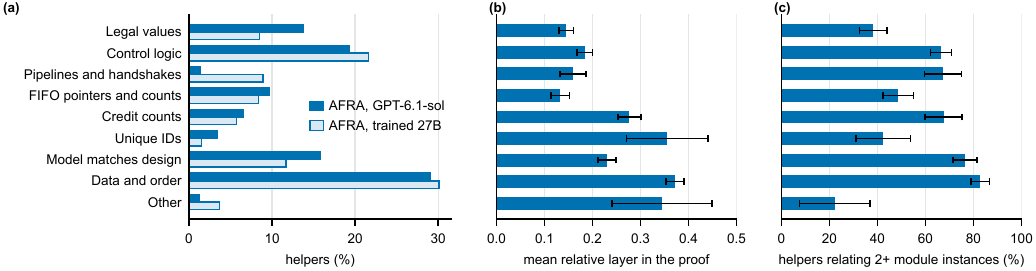}
  \caption{Helper categories. (a) Percentage of the helpers of \afra's closed proofs on the open-source tasks in each
  category, averaged over tasks. (b) Mean relative layer of a helper, its layer divided by the number of layers of its
  proof, in golden proofs of open-source designs. (c) Percentage of the helpers of \afra that relate two or more module
  instances, with error bars for the standard error over tasks in (b) and (c).}
  \label{fig:categories}
\end{figure*}

Fig. \ref{fig:categories}(a) shows the percentage of helpers in each category, averaged over tasks. Data and order is the largest category for both models, nearly a third of the helpers, followed by control logic and model matches design, and few helpers fall into Other. The trained 27B writes more pipeline and handshake helpers, many of which relate several cycles with ``\$past'', while most helpers of GPT-6.1-sol are one-cycle expressions. The stronger model (GPT-6.1-sol) tends to write helpers that better match the quality of those developed by experienced engineers.

In Fig. \ref{fig:categories}(b), we study the correlation between the semantic meaning of a helper and its position in the proof dependency graph. Legal-value, pipeline and FIFO helpers are at the lowest layers, with more than half of them requiring no dependencies. Data-and-order and unique-ID helpers depend more on other helpers, and usually are at higher layers in the graph. In Fig. \ref{fig:categories}(c), we identify the correlation between the helper categories and the number of instances they relate. Most legal-value helpers relate a single instance, while about $80\%$ of the data-and-order helpers relate multiple instances.

\begin{figure}[t]
  \centering
  \includegraphics[width=0.9\columnwidth]{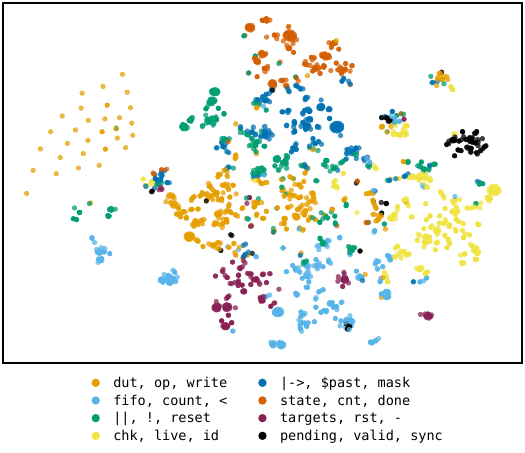}
  \caption{t-SNE of helpers embedded by the tokens of their text, colored by k-means cluster. Each cluster is named by
  its three most over-represented tokens.}
  \label{fig:tsne}
\end{figure}

In Fig. \ref{fig:tsne}, we cluster the helpers with their text tokens. With slang, we parse the operators of each helper, its system functions such as ``\$past'', and its signal names. To focus on the tokens shared across designs, we keep only those that occur in at least 8 tasks and 4 design families. We weigh the tokens with TF-IDF, group the helpers into 8 clusters with k-means, and project them to 2D with t-SNE. We name each cluster by the three tokens its helpers use most often compared with all helpers. Several clusters resemble our categories. For example, the cluster of ``fifo'', ``count'' and ``$<$'' holds the most FIFO helpers, and the cluster of ``state'', ``cnt'' and ``done'' is mostly control logic. We avoid using text-embedding models, which tend to group the helpers by design.

\section{The Complexity of Agentic Proof Closure}
\label{app:proofs}

\begin{figure}[t]
  \centering
  \includegraphics[width=0.9\columnwidth]{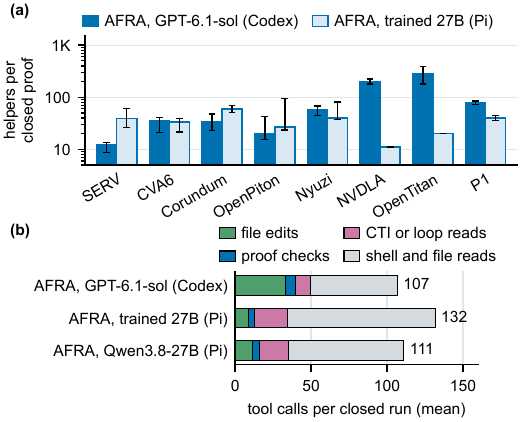}
  \caption{(a) Median and quartiles of helpers per closed proof per family, on the tasks both models close. (b) Mean
  tool calls per closed run, by category.}
  \label{fig:effort}
\end{figure}

We show that constructing a proof is highly non-trivial by analyzing the proof closures of the \afra agents in Table \ref{tab:main} (Fig. \ref{fig:effort}). First, we count the helpers of every closed proof, splitting each helper at its top-level conjunctions and counting the copies of a generate loop once. With GPT-6.1-sol, \afra's constructed proofs have a median of 52 helpers, with 24 proofs having more than 100 helpers. We compare the trained 27B with GPT-6.1-sol by the median and quartiles of helpers in each design family (Fig. \ref{fig:effort}(a)). In general, all the tasks \afra proved need dozens of helpers even after proper accounting (e.g., counting a generate loop once), indicating that it can indeed construct complex proofs.

In Fig. \ref{fig:effort}(b), we group the tool calls of each run that reaches proof closure by category and take the mean over these runs. We show that most of the tool calls go to file reads and tool feedback (CTI/loop) reads, with only a few for file edits and proof checks. Each proof closure requires more than a hundred tool calls on average.

\section{Training Details}
\label{app:train}

\textbf{Training tasks.} We build the training tasks with three agents. A planner agent (GPT-6.1-sol) proposes the tasks, a builder agent (GPT-6.1-sol) builds each task and writes a formal testbench with 12 to 40 candidate targets, and a reviewer agent of another model family (Opus 5.5) reviews each task. We check each target and keep it as a hard target when it stays undetermined after a 4-hour Jasper run, and we cannot prove it with an 8-step induction without helpers within 60 seconds. We admit a training task only when it has no module overlap with a test design, and its 5-gram Jaccard similarity with every test design is below 0.8.

\begin{algorithm}[t]
  \caption{\footnotesize Dependency graph of a golden proof $\mathcal{H}$. $\mathit{CTI}_K(p, S)$: $p$ fails its
  consecution check over $K$ cycles with a CTI, with $S$ assumed.}
  \label{alg:graph}
  \scriptsize
  \begin{algorithmic}[1]
    \Function{FixedPoint}{$S$}
      \Repeat
        \State $F \gets \{p \in S \mid \mathit{CTI}_K(p, S)\}$
        \State $S \gets S \setminus F$
      \Until{$F = \emptyset$}
      \State \Return $S$
    \EndFunction
    \Function{Graph}{$\mathcal{H}$}
      \State $S \gets \Call{FixedPoint}{\mathcal{H}}$, $E \gets \emptyset$
      \For{$h \in S$}
        \State $R \gets S \setminus \{h\}$
        \State $E \gets E \cup \{(h, f) \mid f \in R \setminus \Call{FixedPoint}{R}\}$ \Comment{$f$ depends on $h$}
      \EndFor
      \State \Return $E$
    \EndFunction
  \end{algorithmic}
\end{algorithm}

\textbf{Reward.} Our main idea for progress reward is to calculate the percentage of golden helpers that existing helpers recovered. To do this, we first prove the agent-written helpers with the golden helpers assumed ($H^* \Rightarrow H'$). Then, we assume $P$ and $H'$, and prove the golden helpers ($H' \Rightarrow H^*$). This gives us a proper ``recovery rate'' under two-way implication. To ensure the rigor of the defined potential function, we set $\Phi_{\text{task}}$ to 0 when we reach a proof closure through $H'$.

\textbf{SFT and RFT data.} In SFT, we filter and keep a good Kimi K3 trajectory when its reward exceeds a bar, 0.5 by default. For curriculum tasks, because the reward does not start from zero, we keep a trajectory only when the reward delta before and after agent edits is larger than 0.05. We keep at most 40 trajectories per task and hold out a tenth for validation. In RFT, we apply the same reward filter for good trajectories.

\textbf{Training and serving.} We train with NeMo-RL v0.7.0 at tensor parallelism 8, context parallelism 2 and data parallelism 8, in bf16, keeping an fp32 copy of the weights for the optimizer updates, with Adam ($\beta_1=0.9$, $\beta_2=0.95$, $\epsilon=10^{-8}$), gradient clipping at 1.0, no weight decay, and a cosine learning-rate schedule. We run SFT from $10^{-5}$ to $10^{-6}$, and then RFT for 62 steps from $5\times10^{-6}$ to $5\times10^{-7}$. We serve the models with vLLM 0.30.0.

\section{Extended Training}
\label{app:extended}

After the evaluation of Table \ref{tab:main}, we extend the training by expanding SFT tokens with a slightly expanded set of 128 open-source training designs. 85 of them are from 68 upstream repositories outside of test designs, 2 of them with small edits to be elaborated by our tools, 18 as compositions of open-source modules by an agent, and 3 compiled by XLS~\cite{xls} from existing open-source HLS designs. Another 43 are written by an agent from a high-level specification. We fine-tune the base 27B on 488 Kimi K3 trajectories, a total of 287.4M tokens over the two epochs of SFT, against 167.1M in Sec.~\ref{sec:phi-eval}. We include the summarization calls made when the context is compacted. We then run one RFT round on 47 trajectories of the SFT model, 67 new Kimi K3 trajectories on curriculum tasks and 114 replayed from the SFT set, 133.2M tokens in total over two epochs. With the SFT model, \afra proves 34.2\% ($\pm$3.0) of the targets on average over the 13 families, and with the RFT model 34.8\% ($\pm$3.2), up from 32.7\% of the trained 27B in Table \ref{tab:main}.

\begin{figure}[H]
  \centering
  \includegraphics[width=\columnwidth]{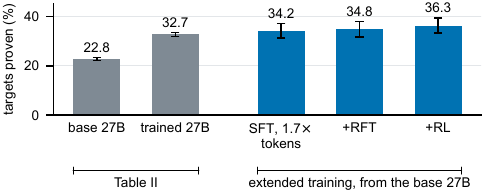}
  \caption{Targets proven by \afra with the base and the trained 27B of Table~\ref{tab:main}, and with the base 27B fine-tuned on 1.7$\times$ SFT tokens, plus one RFT round and 10 RL iterations. We report the mean over 13 design families, with std over two repeated runs.}
  \label{fig:extended}
\end{figure}

We then run 10 iterations of RL with GRPO from the RFT model. Each iteration samples 100 groups of 4 trajectories, or 8 for those we classified as hard tasks during an initial profiling. The rollout sampling is done with temperature 1.0, top-p 0.95, top-k 20 and xhigh reasoning, about 55M output tokens per iteration. The advantage is calculated as the reward minus its group's mean, without dividing by the standard deviation, and the learning rate is $10^{-6}$. Since the checkpoint of each iteration generates the rollouts of the next iteration, we profile each checkpoint by the statistics of these rollouts and keep the best checkpoint. With the RL model, \afra proves 36.3\% ($\pm$3.0) of the targets, as summarized in Fig. \ref{fig:extended}.

We show that scaling reward-filtered high-quality SFT can indeed boost the solve rate by a large margin, potentially due to cloning high-quality reasoning patterns, in which the agent typically reasons over the cycle-by-cycle transitions and the reachable and unreachable states of the design. High accuracy and efficiency in such reasoning are important to success. An extra round of RFT and RL, though effective, only offers marginal improvement, mainly by correcting tool-call errors, pruning overlong thinking and discouraging empty submissions. Although the harness already reminds the agent near the end of a trajectory, we find that the agent still makes empty submissions sometimes. The RL training has not converged yet, and further improvement can come from more RL iterations or better training curriculum and recipe, which we leave to future work.